\documentclass{article}
\usepackage{spconf,amsmath,graphicx}
\usepackage{enumitem}
\setlist{nosep, leftmargin=14pt}
\usepackage{subcaption}
\usepackage{amssymb}
\usepackage[misc]{ifsym}
\usepackage{pifont}
\usepackage{hyperref}
\newlength{\bibitemsep}
\newlength{\bibparskip}
\let\oldthebibliography\thebibliography
\renewcommand\thebibliography[1]{%
  \oldthebibliography{#1}%
  \setlength{\parskip}{\bibitemsep}%
  \setlength{\itemsep}{\bibparskip}%
}
\usepackage{url}
\usepackage{breakurl}

\usepackage[table]{xcolor}
\usepackage{tikz}
\definecolor{maskDDPM}{RGB}{215,211,235}
\definecolor{maskDDPM_border}{RGB}{0,0,0}

\definecolor{syncdr}{RGB}{176, 196, 177}
\definecolor{syncdr_border}{RGB}{0,0,0}

\definecolor{neus}{RGB}{181, 204, 217}
\definecolor{neus_border}{RGB}{0,0,0}

\definecolor{slice}{RGB}{217,232,251}
\definecolor{slice_border}{RGB}{0,0,0}

\definecolor{3d_only}{RGB}{242,232,232}
\definecolor{3d_only_border}{RGB}{242,215,212}

\definecolor{stable}{RGB}{222,222,183}
\definecolor{stable_border}{RGB}{0,0,0}

\definecolor{population}{RGB}{247,214,182}
\definecolor{population_border}{RGB}{0,0,0}

\DeclareRobustCommand{\colorsquare}[1]{\tikz{\path[draw=#1_border,fill=#1, thick, rounded corners=0.6pt] (0,0) rectangle (6pt,6pt);}}

\title{Cascaded Diffusion Models for 2D and 3D Microscopy Image Synthesis to Enhance Cell Segmentation}
\name{Rüveyda Yilmaz (\ding{41}), Kaan Keven, Yuli Wu, Johannes Stegmaier}
\address{Institute of Imaging and Computer Vision, RWTH Aachen University, Germany \\ Email: {\textit{rueveyda.yilmaz@lfb.rwth-aachen.de}}}

\begin{document}
%
\maketitle
\begin{abstract}
Automated cell segmentation in microscopy images is essential for biomedical research, yet conventional methods are labor-intensive and prone to error.
While deep learning-based approaches have proven effective, they often require large annotated datasets, which are scarce due to the challenges of manual annotation.
To overcome this, we propose a novel framework for synthesizing densely annotated 2D and 3D cell microscopy images using cascaded diffusion models.
Our method synthesizes 2D and 3D cell masks from sparse 2D annotations using multi-level diffusion models and NeuS, a 3D surface reconstruction approach.
Following that, a pretrained 2D Stable Diffusion model is finetuned to generate realistic cell textures and the final outputs are combined to form cell populations.
We show that training a segmentation model with a combination of our synthetic data and real data improves cell segmentation performance by up to 9\% across multiple datasets.
Additionally, the FID scores indicate that the synthetic data closely resembles real data.
The code for our proposed approach will be available at \href{https://github.com/ruveydayilmaz0/cascaded_diffusion}{https://github.com/ruveydayilmaz0/cascaded\_diffusion}.
\end{abstract}
\section{Introduction}
\label{sec:intro}
Automatic cell segmentation in microscopy images enables the quantitative analysis of cellular structures and processes, advancing biomedical research in domains such as cancer diagnosis \cite{wang2016automatic}, drug discovery \cite{krentzel2023deep}, and pathological assessment \cite{albayrak2019automatic}.
Conventional methods, such as manual annotation or classical image processing techniques, are time-consuming, prone to human error, and often struggle with the complexity and variability of biological data \cite{liu2021survey}.
Deep learning-based methods, on the other hand, have demonstrated the ability to automatically extract detailed features from images and accurately delineate individual cells, even in complex and noisy environments \cite{albayrak2019automatic, krentzel2023deep, wang2016automatic}.
Despite their success, they generally require large annotated microscopy data for training which is usually scarce due to the difficult manual annotation.
One approach to solve this problem is to use synthetically generated data, which can supplement real datasets and provide diverse, labeled examples for training, improving model generalization and performance \cite{liu2021survey}.
To address this, \cite{related1} introduces a synthetic image generation method for fluorescence microscopy images.
The authors assume convex shapes for real cells and generate random 3D ellipsoids to represent cell volumes, which are subsequently used as annotation labels for the synthetic dataset.
To generate the final images conditioned on these randomly created volumes, they propose an architecture called SpCycleGAN, an extension of CycleGAN \cite{cyclegan}.
Chen \textit{et al.} criticize the use of ellipsoidal cell nuclei volumes, arguing that this approach is inadequate for representing non-convex nuclei \cite{related2}.
As an alternative, they propose employing Bézier curves to create synthetic nuclei shapes.
Constrained by these shapes, they generate the final volumes using the SpCycleGAN architecture \cite{related1}.
Similarly, \cite{eschweiler2024denoising} adopts an approach akin to \cite{related1} for simulating nuclei shapes, but utilizes denoising diffusion probabilistic models (DDPMs) \cite{ddpm} to overlay texture onto the generated shapes.
In addition to generating 3D cell microscopy images, the proposed model also synthesizes 2D videos of living cells modifying the UNet architecture from the DDPM.
\begin{figure*}[t]
\centering
\includegraphics[width=0.85\textwidth]{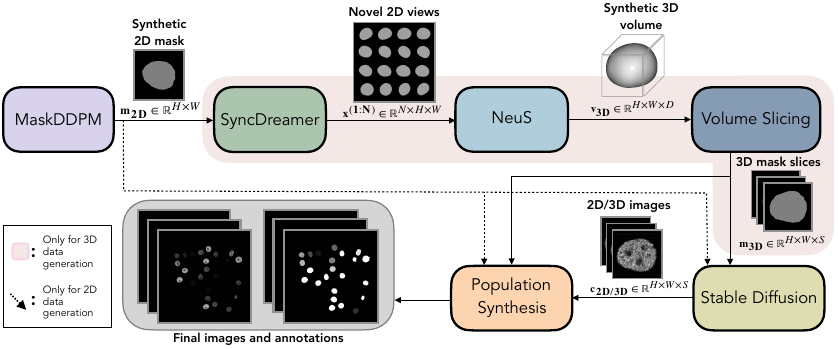}
\caption{Overview of the proposed method. For 2D data synthesis, MaskDDPM (\colorsquare{maskDDPM}) and Stable Diffusion (\colorsquare{stable}) generate masks and cell textures respectively. For 3D data generation, SyncDreamer, NeuS and volume slicing (\colorsquare{3d_only}) are additionally employed. The final images and the masks are combined using the population synthesis module (\colorsquare{population}) .}
\label{fig:main_figure}
\end{figure*}
Alternatively, \cite{yilmaz2024annotated} presents an approach for generating 2D cell microscopy image sequences by employing a 2D DDPM alongside a flow prediction model to produce temporally consistent synthetic sequences.
While the approaches mentioned above offer methods to generate synthetic cell microscopy data in 2D or 3D, they either rely on assumed cell shapes to generate masks or base the masks on real data statistics.
The former approach does not incorporate actual information from real cell shapes, which can lead to unrealistic cell structures due to the imposed assumptions.
The latter approach utilizes real nuclei shape information; however, many 3D datasets lack comprehensive 3D annotations, with labels typically available only for a single z-slice \cite{isbi_challenge}.
This limitation makes it infeasible to train models on fully-annotated 3D data to generate synthetic 3D masks.
Moreover, \cite{related2, eschweiler2024denoising, related1, yilmaz2024annotated} train generative models from scratch to synthesize cell textures.
Yet, the performance could potentially be enhanced by finetuning a pretrained foundation model with the limited data available \cite{controlnet}.
In this paper, we propose an approach for synthesizing densely annotated 2D and 3D cell microscopy images with a cascade of diffusion models and volume reconstruction.
Specifically, we generate synthetic 3D masks from sparse 2D real annotations, introduce a method to finetune the pretrained 2D Stable Diffusion model for 2D and 3D cell texture synthesis and demonstrate that segmentation performance can be improved using the synthetic data produced by the proposed methodology.
\section{Methods}
\label{methods}
\textbf{Mask Synthesis in 2D.}
To define the cell outlines ($\mathbf{m_{2D}} \in \mathbb{R}^{H \times W}$), we start by generating synthetic masks using a limited number of ground truth annotations from real datasets.
Accordingly, we employ a DDPM architecture, which we name MaskDDPM  \mbox{(Fig. \ref{fig:main_figure}, \colorsquare{maskDDPM} )}, as an initial step towards 2D and 3D mask synthesis \cite{ddim}.
In many 3D microscopy datasets, only a single slice in the z-dimension is sparsely annotated \cite{isbi_challenge}, making it impractical to train a 3D model due to the lack of complete cell shape information.
Thus, we also use a 2D MaskDDPM architecture for 3D datasets by training it on the sparse 2D annotations.
Although the output is always 2D, this initial step serves as a foundation for both 2D and 3D image synthesis. \\
\textbf{Multiview-consistent Mask Generation.}
For generating 3D synthetic data, we utilize SyncDreamer \mbox{(Fig. \ref{fig:main_figure}, \colorsquare{syncdr})} to predict novel 2D views ($\mathbf{x^{(1:N)}} \in \mathbb{R}^{N \times H \times W}$) from the 2D output mask $\mathbf{m_{2D}}$ produced by MaskDDPM \cite{syncdreamer}.
SyncDreamer, conditioned on a single 2D image of an object, generates multiple unseen views simultaneously using a DDPM architecture.
To ensure multiview consistency, noise across multiple views is jointly predicted according to the following DDPM training objective \cite{syncdreamer}:
\begin{flalign}
\label{eqn:syncdr}
L\left(\theta\right)=\mathbb{E}_{t, \mathbf{x}_0^{(1: N)}, n, \epsilon^{(1: N)}}\left[\left\|\epsilon^{(n)}-\epsilon_\theta^{(n)}\left(\mathbf{x}_t^{(1: N)}, t\right)\right\|_2\right],
\end{flalign}
\begin{minipage}{\columnwidth}where $\mathbf{x}_t$ is the noisy image at diffusion timestep $t$, $N$ is the number of predicted 2D views, $\epsilon^{(n)}$ and $\epsilon_\theta^{(n)}$ are the added and the predicted noises for the view $n$, respectively.
\end{minipage}\\

\noindent The model is designed to be trained on random 2D views of 3D volumes.
However, as previously mentioned, dense 3D masks are often unavailable in real microscopy sequences.
To address this, we propose training SyncDreamer on spherical harmonic shapes \cite{muller2006spherical}.
Specifically, we generate spherical harmonic volumes and extract multiple 2D surface images from random viewing directions for each volume.
These 2D images form a training dataset that enables SyncDreamer to learn how to predict additional views of a shape from a single 2D view.
During inference, conditioned on $\mathbf{m_{2D}}$, the output from MaskDDPM, SyncDreamer predicts multiple novel views for the corresponding hypothetical 3D volume. \\
\textbf{Multiview-image to Volume Generation.}
To generate a dense volumetric mask ($\mathbf{v_{3D}} \in \mathbb{R}^{H \times W \times D}$) from the predicted multiple views, we employ NeuS \mbox{(\cite{wang2021neus}, Fig. \ref{fig:main_figure} \colorsquare{neus})}.
NeuS is a surface reconstruction method that represents object surfaces as signed distance functions, modeled by multi-layer perceptrons (MLPs).
The MLPs are trained to align the rendered surface images with the provided multiview inputs.
Using the multiview 2D mask predictions $\mathbf{x^{(1:N)}}$ from SyncDreamer, we construct 3D volumes $\mathbf{v_{3D}}$ via NeuS.
Consequently, rather than relying on randomly generated spheres as masks for synthetic data \cite{eschweiler20213d, related1, cell_mask_classical2}, our approach transfers information from sparse 2D ground truth annotations into the synthetic data generation process.\\
\textbf{Slicing the Synthetic Masks.}
Since real 3D cell microscopy images typically consist of multiple 2D slices representing a 3D volume along the z-dimension \cite{isbi_challenge}, we slice the output volumes $\mathbf{v_{3D}}$ from NeuS at equal intervals ($\mathbf{m_{3D}} \in \mathbb{R}^{H \times W \times S}$, where \textit{S} is the number of slices) to align their structure with that of the real datasets \mbox{(Fig. \ref{fig:main_figure}, \colorsquare{slice})}.
\setlength{\abovedisplayskip}{0.2pt}
\begin{figure}[htb]
\centering
\includegraphics[width=0.40\textwidth]{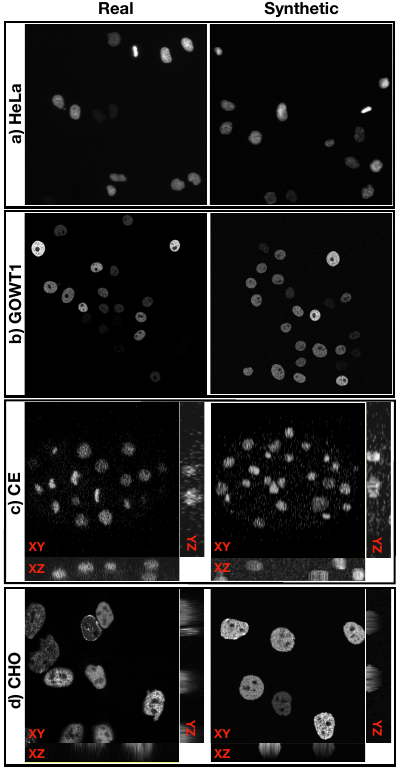}
\caption{Representative examples from the real and synthetic (a) HeLa, (b) GOWT1, (c) CE, and (d) CHO datasets. For the 3D CE and CHO datasets, zoomed-in 2D cross-sectional images along the X and Y dimensions are also provided.
}
\label{qualitative}
\end{figure} \\
\textbf{Cell Texture Generation.}
To apply realistic cell textures ($\mathbf{c_{2D/3D}} \in \mathbb{R}^{H \times W \times S}$) to the shapes defined by the synthetic masks $\mathbf{m_{2D}}$ or $\mathbf{m_{3D}}$, we utilize Stable Diffusion, a foundational text-to-image diffusion model \cite{ldm}. 
For microscopy image generation, it is crucial that the cell shapes in the generated images accurately correspond to the synthetic masks, as these image-mask pairs are usually intended for training segmentation models.
To ensure this alignment, we employ a modified version of the Stable Diffusion architecture \mbox{(\cite{controlnet}, Fig. \ref{fig:main_figure} \colorsquare{stable})}.
Specifically, instead of using text prompts, we condition the model on synthetic masks and allow it to overlay realistic cell textures. This is achieved by copying the UNet weights from the pretrained Stable Diffusion model and connecting it to the image-based conditional input via convolutional blocks at multiple scales.
During training, these additional convolutional blocks are learned, while the pretrained UNet weights from Stable Diffusion are finetuned using real cell microscopy images and their ground truth annotations as shape-conditioned inputs.\\
\indent Stable Diffusion is designed for generating 2D outputs, but not for 3D data generation.
To harness its capabilities for 3D image synthesis, we propose a method that enables the use of the 2D model for 3D data generation.
We generate images for each slice of the 2D synthetic masks using the mask-conditional Stable Diffusion model that we finetuned. However, generating 2D slice images independently would disrupt the continuity of textures across slices for each cell.
To preserve this continuity, we exploit the inherent properties of Stable Diffusion, which operates as a Denoising Diffusion Implicit Model (DDIM) \cite{ddim} when the hyperparameter $\eta$, controlling the stochasticity of the diffusion process, is set to 0.
In DDIMs, changes in the latent noise space are deterministically reflected in the image space, as randomness is eliminated by setting $\eta=0$.
Through this property, the latents $\mathbf{c}_{T}^0,...,\mathbf{c}_{T}^S$ can be manipulated to have the desired amount of correlation in synthetic images $\mathbf{c}_{0}^0,...,\mathbf{c}_{0}^S$.
Correspondingly, inspired by \cite{ge2023preserve}, we set a common noise vector $\mathbf{c}_{cmn}$ that is shared among all the slices of each 3D image, and a unique noise vector $\mathbf{c}_{unq}^s$ that is independently generated for each slice.
Combinations of these noise vectors are used to generate the latent noise samples $\mathbf{c}_{T}^0,...,\mathbf{c}_{T}^S$ as follows:
\begin{flalign}
\label{eqn:common_noise}
\mathbf{c}_{cmn}, \mathbf{c}_{unq}^s \sim \mathcal{N}\left(\mathbf{0}, \mathbf{I}\right), \qquad \notag \\ \mathbf{c}_{T}^s=\frac{\rho}{\sqrt{1+\rho^2}}\cdot \mathbf{c}_{cmn}+ \frac{1}{\sqrt{1+\rho^2}} \cdot \mathbf{c}_{unq}^s\quad,
\end{flalign}
\begin{minipage}{\columnwidth}where $s$ is the slice number and $\rho$ is the strength of the texture consistency among slices. The noise vectors are scaled by constants dependent on $\rho$ such that $\mathbf{c}_{T}^s$ has unit variance.
\end{minipage}\\

\noindent Using the approach outlined in Equation \ref{eqn:common_noise}, the slices from each synthetic image $\mathbf{c_{3D}}$ exhibit structural similarity through $\mathbf{c}_{cmn}$, while also incorporating some variation through $\mathbf{c}_{unq}^s$. \\
\textbf{Cell Population Synthesis.}
As the process of data synthesis described thus far involves one cell per image, we ultimately merge the outputs to create cell populations. 
Following the approach proposed in \cite{cell_mask_classical2}, we initialize two empty canvases for each synthetic raw image-mask pair ($\mathbf{c_{2D/3D}}, \mathbf{m_{2D/3D}}$).
Random synthetic cells are iteratively placed on the raw image canvas at unoccupied locations, with the corresponding masks aligned at the same positions on the mask canvas.
To simulate the clustering of cells as observed in real datasets, each new cell is positioned near the previously added cells with a certain probability.
This probability is modeled using a Bernoulli distribution with parameter $p \in [0,1]$, referred to as the \textit{clustering probability}.
\begin{table*}[tbp]
\caption{Qualitative results for (a) the segmentation performance (SEG), (b) FID, and (c) ablation experiments.}
\centering 
\begin{subtable}{0.4\textwidth}
    \centering 
    \begin{tabular}{lcccc}
        \hline 
        Dataset & None & Real & Synthetic & Both \\
        \hline 
        HeLa & 0.775 &  0.826 & 0.869 & \textbf{0.877} \\
        GOWT1 & 0.471 & 0.881 & 0.905 & \textbf{0.914} \\
        CHO & 0.698 & 0.897 & 0.893 & \textbf{0.909} \\
        CE & 0.212 & 0.706 & 0.738 & \textbf{0.793} \\
        \hline
    \end{tabular}
    \vspace{0.5cm}
    \caption{The segmentation performance (SEG $\uparrow$) of Cellpose generalist model without finetuning (None), with finetuning on real, synthetic, and both datasets.}
    \label{table:comparisons}
\end{subtable}%
\hspace{0.2cm} 
\begin{subtable}{0.2\textwidth}
    \centering
    \begin{tabular}{cc}
        \hline 
        $\mathrm{FID_{r2r}}$ & $\mathrm{FID_{r2s}}$ \\
        \hline 
        5.7 &  6.1 \\
        0.9 & 3.4 \\
        0.3 & 0.7 \\
        1.2 & 2.5 \\
        \hline
    \end{tabular}
    \vspace{0.5cm}
    \caption{The FID score ($\downarrow$) between two sets of real data and between real and synthetic data.}
    \label{table:FIDs}
\end{subtable}%
\hspace{0.2cm} 
\begin{subtable}{0.25\textwidth}
    \centering
    \begin{tabular}{cc}
        \hline 
        w/o SyncDr. & w/o pretr. \\
        \hline 
        - & 0.843 \\
        - & 0.889 \\
        0.876 & 0.890 \\
        0.788 & 0.791 \\
        \hline
    \end{tabular}
    \vspace{0.5cm}
    \caption{The segmentation performance (SEG $\uparrow$) of Cellpose in ablation experiments without SyncDreamer (w/o SyncDr.) or pretrained Stable Diffusion (w/o pretr.).}
    \label{table:ablation}
\end{subtable}
\vspace{-2em}
\end{table*}

\section{Experiments}
\label{sec:pagestyle}
\textbf{Datasets.}
To demonstrate the proposed methodology, we experiment with four public cell nuclei datasets made available by the Cell Tracking Challenge \cite{isbi_challenge}.
Each dataset comprises two grayscale confocal microscopy image sequences, with lengths ranging from 92 to 250 frames.
These datasets include 2D images of HeLa cancer cells expressing the H2B-GFP protein and mouse stem cells expressing GFP-Oct4, 3D images of Chinese hamster ovary cells expressing GFP-PCNA and developing embryo cells of \textit{C. elegans} expressing H2B-GFP.
For convenience, we abbreviate them as HeLa, GOWT1, CHO, and CE, respectively, throughout this paper.
The datasets include manually annotated sparse gold truth labels and densely annotated silver truth labels, which are obtained through a voting mechanism among multiple cell segmentation models \cite{isbi_challenge}.
In each dataset, we use one of the annotated sequences for training and the other for testing.
Before training, we identify the cells in the images using the silver truth masks and create crops of size 128$\times$128 for both the cells and their corresponding silver truth masks.
This step is essential, as SyncDreamer cannot predict multiview images when multiple masks are present in the same frame.
The generated mask crops are subsequently used for training MaskDDPM, while the image-mask pairs are employed for finetuning Stable Diffusion.
\\
\textbf{Results.}
To evaluate the quality of our synthetic data, we experiment with a cell segmentation method called Cellpose \cite{stringer2021cellpose}.
As Cellpose is a generalist model designed to segment unseen data, we first assess its performance on one of the annotated real sequences, which we reserve solely for testing.
Next, we finetune it on the remaining real sequence, on 1000 synthetic images, and on the mixture of both real and synthetic data.
As a result, we have four Cellpose models trained with different combinations of data for each of the four datasets.
To assess the performance of each model, we compute SEG, which calculates the intersection over union (IoU) between the predicted and the ground truth segmentation masks \cite{isbi_challenge}.
The results given in Table \ref{table:comparisons} indicate that Cellpose does not consistently perform well without finetuning, particularly for the GOWT1 and CE datasets.
Notably, finetuning the model with either real or synthetic data significantly enhances the performance, with the best results achieved using both. \\
\indent As another evaluation metric, we compute the Fréchet Inception Distance (FID) on real and synthetic data \cite{fid}.
First, to establish a baseline for the typical distance between two sets of real data, we calculate the FID between the real training and the test sets and name this $\mathrm{FID_{r2r}}$.
Next, we calculate the FID between the real test data and 1000 synthetic data, and denote this as $\mathrm{FID_{r2s}}$.
As shown in Table \ref{table:FIDs}, $\mathrm{FID_{r2r}}$ and $\mathrm{FID_{r2s}}$ are sufficiently close, suggesting that the synthetic data can be considered realistic, as discussed in \cite{dhariwal2021diffusion}. 
In Fig. \ref{qualitative}, we present representative example images from the real and synthetic datasets. \\
\textbf{Ablation experiments.}
To assess the contribution of our proposed 3D mask generation approach \mbox{(Fig. \ref{fig:main_figure}, \colorsquare{3d_only} )} and the pretrained Stable Diffusion \mbox{(Fig. \ref{fig:main_figure}, \colorsquare{stable} )} on the overall performance, we conducted ablation experiments by replacing the output from NeuS \mbox{(Fig. \ref{fig:main_figure}, $\mathbf{v_{3D}}$)} by spherical harmonic shapes and Stable Diffusion by a Latent Diffusion Model (LDM) \cite{ldm} without pretraining.
Subsequently, we train Cellpose with a mixture of the resulting synthetic data and the real data. 
As shown in Table \ref{table:ablation}, replacing $\mathbf{v_{3D}}$ while keeping the other components unchanged leads to a reduction in segmentation performance for both the CHO and CE datasets.
This ablation experiment is not applicable to 2D datasets, as the ablated components are not used for 2D image synthesis in the original pipeline.
When an LDM is trained from scratch for the same number of epochs as Stable Diffusion, a comparable decline in performance is observed, and it requires significantly longer training to achieve the same performance as the finetuned Stable Diffusion.
\section{Conclusion}
In this work, we introduce a novel pipeline to synthesize 2D and 3D microscopy images using \textit{ad hoc} diffusion models for generating single-view 2D masks with MaskDDPM, 3D mask volumes with SyncDreamer and NeuS, and cell textures with Stable Diffusion. The resulting 3D volumes are geometrically consistent due to high multiview coherence and both 2D and 3D data are realistic as evidenced by the FID score. Besides, the synthetic data is capable of augmenting real datasets, leading to improved segmentation accuracy across various cell microscopy datasets as the ultimate goal. 
Future work will explore extending the pipeline to different biomedical imaging modalities and applying it to time-series data.

\section{COMPLIANCE WITH ETHICAL STANDARDS}
This research was conducted with public datasets from the Cell Tracking Challenge \cite{isbi_challenge} and ethical approval was not necessary.

\section{Acknowledgments}
This work was supported by the German Research Foundation DFG (STE2802/2-1).

\bibliographystyle{IEEEbib}
\bibliography{main}

\end{document}